\documentclass[preprint,12pt]{elsarticle}
\graphicspath{ {./figures/} }
\usepackage{hyperref}
\usepackage{float}
\usepackage{verbatim} 
\usepackage{apalike}
\usepackage{amssymb}
\usepackage{amsmath}
\restylefloat{figure}
\restylefloat{table}
\usepackage{algorithm}
\usepackage{algpseudocode}
\usepackage{subcaption}
\usepackage[dvipsnames]{xcolor}

\usepackage{booktabs} 
\usepackage{tabularx} 
\usepackage{siunitx} 
\usepackage{mathrsfs}
\usepackage{booktabs,siunitx}

\journal{Expert Systems with Applications}

\biboptions{authoryear}

\begin{document}

\begin{frontmatter}










\title{Uni-3DAD: GAN-Inversion Aided Universal 3D Anomaly Detection on Model-free Products}

\author[label1]{Jiayu Liu}
\ead{liuj35@rpi.edu}

\author[label2]{Shancong Mou}
\ead{mou00006@umn.edu}

\author[label3]{Nathan Gaw}
\ead{Nathan.Gaw@afit.edu}

\author[label1]{Yinan Wang}
\ead{wangy88@rpi.edu}

\cortext[cor1]{Corresponding author: Yinan Wang}
\address[label1]{Department of Industrial and Systems Engineering, Rensselaer Polytechnic Institute, Troy, NY, 12180, USA}
\address[label2]{Department of Industrial and Systems
Engineering, University of Minnesota - Twin Cities, Minneapolis, MN, 55455, USA}
\address[label3]{Department of Operational Sciences, Air Force Institute of Technology, OH, 45433, USA}

\begin{abstract}
Anomaly detection is a long-standing challenge in manufacturing systems, aiming to locate surface defects and improve product quality. Traditionally, anomaly detection has relied on human inspectors or image-based methods. However, 3D point clouds have gained attention due to their robustness to environmental factors and their ability to represent geometric data. Existing 3D anomaly detection methods generally fall into two categories. One compares scanned 3D point clouds with design files, assuming these files are always available. However, such assumptions are often violated in many real-world applications where model-free products exist, such as fresh produce (i.e., ``Cookie", ``Bagel", ``Potato", etc.), dentures, bone, etc. The other category compares patches of scanned 3D point clouds with a library of normal patches named memory bank. However, those methods usually fail to detect incomplete shapes, which is a fairly common defect type  (i.e., missing pieces of different products). The main challenge is that, unlike missing regions in images, which manifest as different pixel values or patterns compared to a normal image patch, missing areas in 3D point clouds represent the absence of scanned points. This makes it infeasible to compare the missing region (with no representation in the recorded 3D Scan) with existing point cloud patches in the memory bank. To address these two challenges, we proposed a unified, unsupervised 3D anomaly detection framework capable of identifying all types of defects on model-free products. Our method integrates two detection modules: a feature-based detection module and a reconstruction-based detection module. Feature-based detection covers geometric defects, such as dents, holes, and cracks, while the reconstruction-based method detects missing regions. Additionally, we employ a One-class Support Vector Machine (OCSVM) to fuse the detection results from both modules. The results demonstrate that (1) our proposed method outperforms the state-of-the-art (SOTA) methods in identifying incomplete shapes and (2) it still maintains comparable performance with the SOTA methods in detecting all other types of anomalies.
\end{abstract}


\end{frontmatter}

\section{Introduction}
\label{sec:1}

\begin{figure}[!ht]
\centering\includegraphics[width=1.0\linewidth]{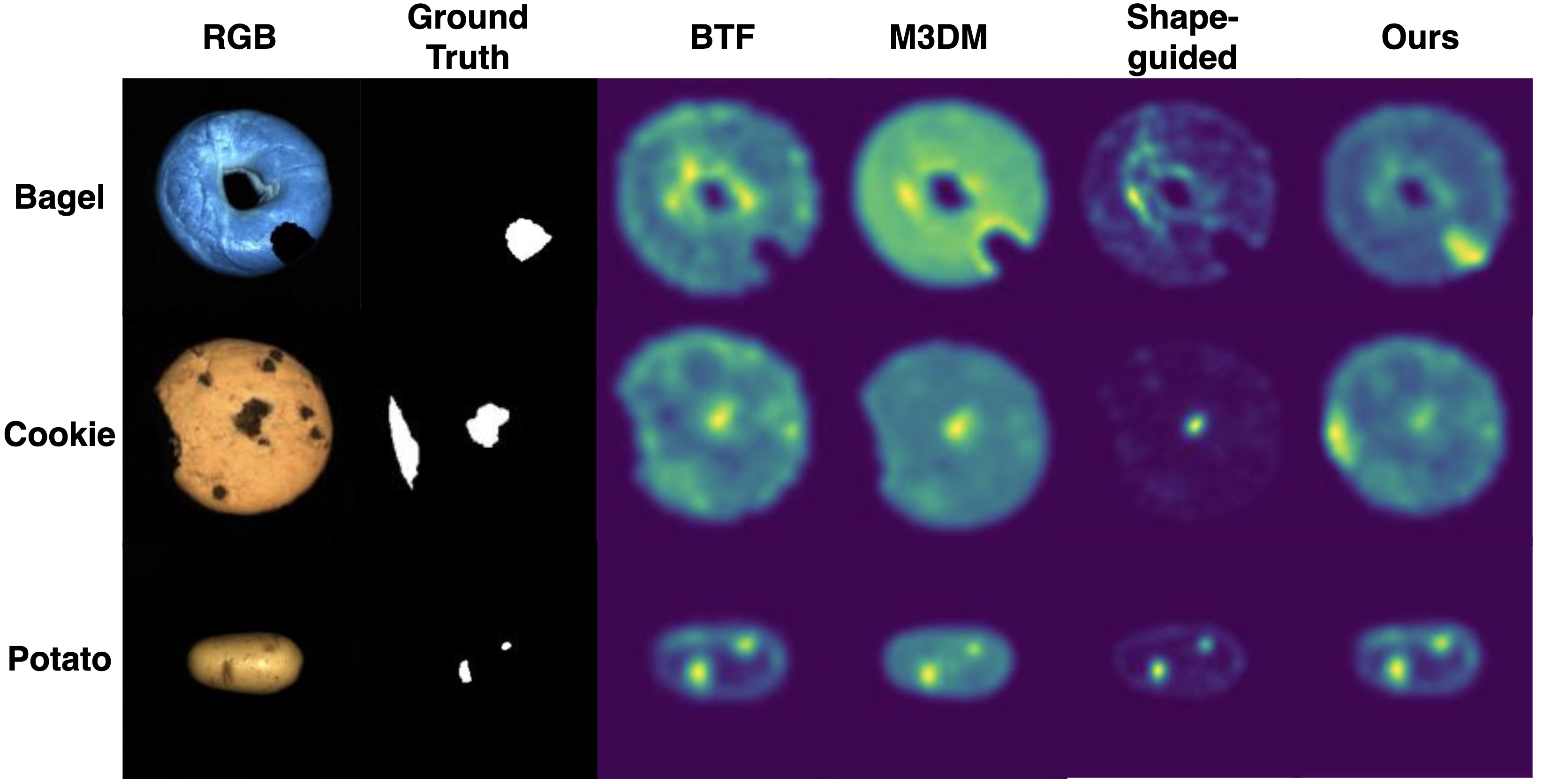}
\caption{Comparison of visual results between our proposed and current SOTA (BTF \citep{horwitz2022empirical}, M3DM \citep{wang2023multimodal}, Shape-guided \citep{shapeguide}) methods on the MVTec 3D-AD dataset \citep{Bergmann_Jin_Sattlegger_Steger_2022b}. Bright areas indicate high anomaly scores. Our method successfully localizes missing regions as defects without compromising the detection accuracy of the other geometric defects, such as dents, holes, and scratches.}
\label{fig:visualresult}
\end{figure}

Surface defects, including cracks, dents, and missing parts, are common during manufacturing and can significantly weaken a component's structural integrity. Anomaly detection is crucial for maintaining high-quality standards by identifying defects that could lead to failures or poor performance. This challenge has been recognized since the beginning of the Industrial Revolution, when anomaly detection was heavily reliant on labor-intensive human visual inspections. However,  human inspectors are historically prone to errors due to carelessness and fatigue. With the emergence of automated defect detection technologies, it has become possible to significantly reduce dependence on human inspection while enhancing efficiency and reliability in industrial quality control. The simplest approach involves directly comparing the pending to-detect samples to a computer-aided design (CAD) file to identify differences. However, this method assumes the assumption that the design file exists, which is not always the case in many industrial settings. Hence,  researchers have focused on developing anomaly detection methods for \textbf{model-free} products, for which the CAD files are unavailable. 

Image-based defect detection techniques on model-free products have significantly advanced.
Several methods have been developed, including filter \citep{fileter1}, tensor-based method \citep{7493677}, and wavelet transformation \citep{7239653}. Recently, with the rise of deep learning, the focus has shifted from traditional approaches like tensor-based decomposition, filters, and digital processing to a range of deep learning methods applied to two-dimensional (2D) RGB images \citep{Li_Zhao_Pan_2017, Cheon_Lee_Kim_Lee_2019a, Wang_Guo_Yue_2021, mou2024paedid}. Although these methods have been proven effective in many industrial applications, they exhibit limitations: specifically, 2D RGB data are less robust than 3D point clouds since (i) they are susceptible to variations in camera angles and lighting conditions, and (ii) they lack the rich geometric information needed for detecting minimal surface defects. 

Addressing the limitations of 2D anomaly detection methods, 3D detection methods have garnered more attention than before. 3D anomaly detection methods typically utilize 3D point clouds, which consist of spatial coordinates that accurately represent the surface topology of products. Compared to 2D images, 3D point clouds offer significant advantages: (i) they maintain the integrity of surface shapes regardless of viewing angles and various lighting conditions, and (ii) they provide comprehensive geometric details crucial for detecting tiny surface defects. Recent state-of-the-art (SOTA) unsupervised 3D anomaly detection methods demonstrate comparable or even better detection results than 2D methods \citep{horwitz2022empirical, shapeguide, wang2023multimodal} on MVTec 3D-AD dataset \citep{Bergmann_Jin_Sattlegger_Steger_2022b}. The core composition of these methods is the 3D point cloud feature extractors that are built on the $k$-nearest-neighbor (KNN) algorithm. By extracting local geometric features from training and testing samples and comparing the differences between them, these methods effectively identify and detect surface defects. We term these methods as \textbf{feature-based methods}. Despite these successes, these feature-based methods primarily focus on identifying defects that can be represented by local geometric features in the point cloud. However, for those defects that cannot be represented by local geometric features, i.e., the product is incomplete and has missing regions, the feature-based method failed to accurately identify them. Conversely, in recent years, with the advent of new 3D anomaly detection datasets \citep{Li_2024_CVPR, liu2023real3d} and fast growth in 3D generative models, \textbf{reconstruction-based methods} have become popular \citep{Li_2024_CVPR,zhou2024r3dad}. The nature of reconstruction-based methods is different from the feature-based methods: they generate the normal sample and compare it with the input sample for defect detection. However, the existing reconstruction-based methods have two limitations: (1) the current 3D generative model showed success in reconstructing the overall shape of the product while failing to fully recover the high-resolution geometric information in local surfaces. Reconstruction-based methods failed to identify defects from surfaces with complex geometric shapes. (2) The generated sample might introduce false alarms to the detection results by over- or under-estimating the shape of the input sample. Therefore, significant research gaps still exist in 3D anomaly detection.

To address the abovementioned issues, this work proposes a unified framework for detecting all types of geometric defects in the 3D point clouds. This architecture fuses two branches of detection methods: feature-based and reconstruction-based. The feature-based branch mainly focuses on surface defects that can be represented by local geometric features, while the reconstruction-based branch targets the detection of missing regions. In addition, Generative Adversarial Network (GAN)-Inversion is first introduced in 3D anomaly detection to ensure the generated sample in the reconstruction-based branch is the normal sample that is most similar to the input sample, reducing the false alarms. Finally, One-Class Support Vector Machine (OCSVM) is tailored to fuse the detection results from two branches and generate the overall anomaly score. The performance of our proposed methods is evaluated on two datasets: (1) the original MVTec 3D-AD dataset and (2) the augmented dataset derived from the original MVTec 3D-AD dataset, which contains extra incomplete samples. 

We summarize our contributions as follows:

\begin{itemize}
    \item Our proposed method has notable adaptability to model-free industrial products.
    \item The two-branch design (feature-based and reconstruction-based branches) enables the model to detect all kinds of defects. (capable of identifying incomplete shapes). 
    \item The model maintains the unsupervised nature that only uses non-defective unlabeled training data, which eases the burden on point-wise labeling.
\end{itemize}

The rest of the paper is organized as follows: Sec. \ref{sec:2} provides a concise overview of related works in anomaly detection. Sec. \ref{sec:3} introduces our proposed method, including each branch's
details. Sec. \ref{sec:4} presents the experimental design, results, and a discussion of an ablation study. Finally, Sec. \ref{sec:5} summarizes the main findings and concludes the work.

\section{Related Works}\label{sec:2}
\label{Related Works}
In industrial anomaly detection, there are three main research areas, including (i) few-shot anomaly detection, in which limited data with pixel/point-wise labels are available for training; (ii) unsupervised anomaly detection, which aims to reduce human labor in labeling; and (iii) noisy anomaly detection, in which the data is often corrupted. Our research primarily focuses on unsupervised anomaly detection methods, which aligns well with the practical scenario where pixel/point-wise labels are usually unavailable. We first review popular 2D anomaly detection methods and then extend the review to include 3D anomaly detection approaches. 

\subsection{2D Anomaly Detection}
\cite{Bergmann_Batzner_Fauser_Sattlegger_Steger_2021} released an industrial dataset for unsupervised anomaly detection, which has become prevalent in recent 2D anomaly detection research. Additionally, \cite{Benchmarking} summarizes two mainstream unsupervised 2D deep learning anomaly detection methods: (i) reconstruction-based methods and (ii) feature-based methods. 

\subsubsection{2D Reconstruction-Based Methods}
Reconstruction-based methods represent a significant strand in 2D anomaly detection. These methods are grounded in generative models, which are trained to generate normal samples. Therefore, when the testing sample comes in, the generative model will generate the most similar normal sample to the testing sample. The defects are then detected by comparing the testing sample with the generated normal samples. A notable approach of these methods leverages auto-encoders for image reconstruction and subsequent anomaly region identification, as demonstrated in studies \citep{autoBergmann, Zavrtanikauto, Zavrtanik_auto}.
Additionally, there is a growing interest in employing GAN-based methods on 2D anomaly detection \citep{anogan, RGI}, since GAN can better capture the data distribution and hence generate high-quality data. However, it is difficult to train the GAN network. The most recent advancements in this domain have been directed towards exploring the potential of diffusion models \citep{song2021scorebased}, an emerging technique showing promise and easy training compared with the GAN, particularly in medical anomaly detection. \cite{anoddp} further exemplified this direction by proposing a novel anomaly detection framework utilizing diffusion models.

On the other hand, normalizing flows, which are another popular generative model, can perform 2D anomaly detection as well \citep{normalizingliu, Rudcsflow}. Unlike GAN-based detection methods, normalizing flows converts the normal input images into multi-variable Gaussian distribution and learns the converting process. When performing anomaly detection, the model aims to separate the in-distribution (normal) and out-of-distribution (abnormal) samples with the trained model. CFLOW-AD is a conditional normalizing flow that performs the positional encoding on the input 2D images \citep{csflow_ad}. The model is faster and smaller than the other normalizing flow models. \cite{csflow} use multi-scale features jointly to estimate the distribution throughout normalizing flow blocks for 2D anomaly detection.

\subsubsection{2D Feature-based Methods}
Feature-based methods use different feature extractors to obtain representative features from normal images. After feature extraction, these features can be stored as discrete reference feature sets (referred to as the memory bank). Then, when the testing sample comes in, the same feature extractor is applied, and the difference between the testing features and features in the memory bank is used to calculate the anomaly score based on the predefined distance metric, thereby localizing the anomaly regions. 

SPADE adopts $K$ nearest features (in the memory bank) to the testing feature to calculate the anomaly score and detect defects \citep{SPADE}. \cite{patchcore} addresses the problem of long inference time in SPADE and proposes using a subset of the memory bank to increase computation speed. \cite{PADIM} first proposes a continuous representation of the reference feature set by fitting a multi-variate Gaussian distribution. \cite{FAPM} uses a patch-wise memory bank, which largely reduces the inference time during anomaly detection. 

The teacher-student network is another popular feature-based anomaly detection technique. 
The teacher-student network anomaly detection method usually consists of two identical networks: the teacher network (pre-trained on a large dataset) and the student network (trained against the output of the teacher network on the anomaly-free dataset). During testing, the defects are marked when the outputs from the teacher network and the student network in the same area are different.
\cite{2Dts} is the first to utilize teacher-student structure on 2D anomaly detection. In addition, \cite{vggst} and \cite{stf} use a more compact and powerful network and multi-scale features for detection. \cite{ast} adopt an asymmetric teacher-student network; namely, the teacher network has a different structure from the student network. This structure can increase the gap between the output of the teacher network and the student network on the testing data, which is beneficial for anomaly detection.


\subsection{3D Anomaly Detection}
3D anomaly detection is different from 2D anomaly detection because of the nature of the dataset. The current popular 3D point cloud dataset for industrial 3D anomaly detection is MVTec 3D-AD \citep{Bergmann_Jin_Sattlegger_Steger_2022b}. The dataset contains RGB images and 3D point clouds of both model-free and model-fixed samples. In recent two years, with the development of 3D sensors, there have been several newly released 3D datasets for the anomaly detection task \citep{liu2023real3d, Li_2024_CVPR}. Compared with the bloom of 2D anomaly detection, the methods of 3D anomaly detection are limited. Despite limited methods, there are two main branches for 3D anomaly detection: feature-based methods and reconstruction-based methods. 

\subsubsection{3D Reconstruction-based Methods}
Similar to 2D reconstruction-based methods, 3D reconstruction-based methods also have generative models that are trained on the normal samples. Voxel f-AnoGAN, which utilizes 3D voxel grids, performs anomaly detection in the 3D brain data, opening a new direction for 3D reconstruction-based methods in the medical field \citep{fano}. This represents a significant expansion of the 2D GAN-based anomaly detection method \citep{anogan}. Historically, the development of 3D reconstruction-based methods in industrial settings has been hindered by the absence of appropriate datasets for 3D generative models. This situation changed with a new synthetic 3D anomaly detection dataset, Anomaly-ShapeNet, which has significantly propelled the 3D reconstruction-based methods forward \citep{Li_2024_CVPR}. In their work, they introduce IMRNet, a self-supervised learning network that utilizes a masking and reconstruction strategy for detecting 3D anomalies. Furthermore, R3D-AD employs the 3D diffusion model to detect defects \citep{zhou2024r3dad}. However, they still face the challenge that the generated 3D point clouds only capture the shape of the target products, indicating they are inadequate for reconstructing the high-resolution 3D point clouds for surfaces with complex geometric features. This limitation is primarily caused by the characteristics of the training dataset, the capacity of generative models, and the nature of the loss functions in training. The dataset built upon ShapeNet \citep{chang2015shapenet} has around 2000 points in each sample of the 3D point cloud, which is insufficient for high-resolution surface representation. In addition, the 3D point cloud with high-resolution surface representation typically contains millions of points to represent a whole product, which exceeds the capacity of existing generative models. Furthermore, commonly used loss functions, such as Mean Square Error (MSE) and Earth Mover's Distance (EMD), tend to prioritize the accuracy of the overall shape over local surface details.



\subsubsection{3D Feature-based Methods}

Inspired by the success of 2D teacher-student anomaly detection, \cite{Bergmann_Sattlegger_2023} first introduced teacher-student networks to detect defects from 3D point clouds. In addition, \cite{ast} utilizes the depth image instead of the 3D point clouds in the asymmetric teacher-student network for anomaly detection. Although these methods have successfully identified surface defects in 3D point clouds, they are constrained by the need for additional training datasets or suffer from reduced accuracy in localization since there is little difference between the output of the teacher network and that of the student network. Meanwhile, a different line of research has explored using memory banks for 3D anomaly detection \citep{horwitz2022empirical, wang2023multimodal, shapeguide, cao2023CPMF}. These studies have shown that memory bank-based methods perform effectively on objects with complex geometric shapes. However, these approaches still have significant limitations. One issue is the large computational complexity since these methods require the calculation of a pairwise distance matrix between the features in the memory bank and those of samples. This implies that the computational cost largely depends on the size of the memory bank, making it impractical for online detection. Moreover, these methods typically rely on both well-aligned 2D images and 3D point clouds for anomaly detection, which may not be practical in real-world industrial environments where 2D images (and/or the alignment) are not always available. 

However, without the aid of 2D images, incomplete shape detection --  a fairly important defect detection task in industrial applications -- using current 3D feature-based methods is highly challenging when relying solely on 3D scan data. The primary difficulty lies in the fact that unlike missing regions in images—where differences are reflected in pixel values or patterns compared to a normal product image—missing areas in 3D point clouds are characterized by the absence of scanned points. This absence makes it impossible to compare the missing region with existing point cloud patches in the memory bank. To address these limitations, we propose a unified 3D anomaly detection model to fuse both feature-based and reconstruction-based methods for the industrial products anomaly detection task.

\section{GAN-Inversion Aided Universal 3D Anomaly Detection}\label{sec:3}
\subsection{Overview}
\begin{figure}[!ht]
\centering\includegraphics[width=1.0\linewidth]{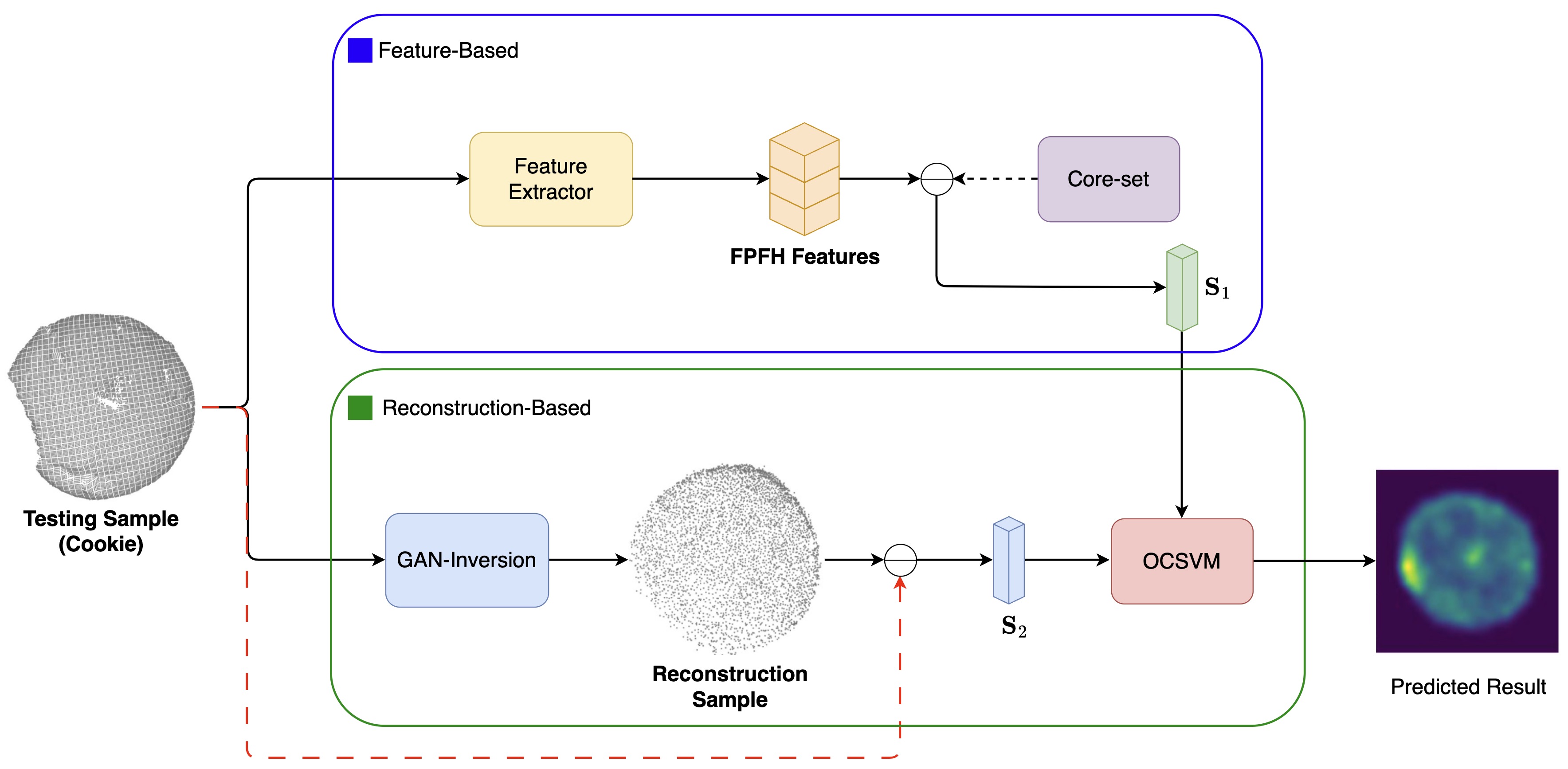}
\caption{Inference of our proposed method. Our proposed method has three basic modules: (i) feature-based module, (ii) reconstruction-based module, and (iii) OCSVM fusion module. The design of the proposed method can facilitate the detection of universal defects.}
\label{fig:pipeline}
\end{figure}

Our proposed model is comprised of two key branches and one fusion module: a feature-based branch (Sec. \ref{sec:feature}), a reconstruction-based branch (Sec. \ref{Sec:recon}), and a fusion module for the final decision (Sec. \ref{sec:OCSVM}). The feature-based module captures the local geometric information of the products, enabling the identification of defects on the local surfaces (e.g., dents, cracks, etc.). In contrast, the reconstruction-based method focuses on localizing and restoring the overall shape of the products by extracting global geometric features, which is a key function that the feature-based method fails to perform. Finally, we use an OCSVM to fuse the outputs of these two branches and output the final anomaly score \citep{ocsvm}. The architecture of our proposed method is illustrated in Fig. \ref{fig:pipeline}.

\subsection{Feature-based Anomaly Detection}\label{sec:feature}
The first component of our proposed method is feature-based anomaly detection. As discussed in Sec. \ref{sec:1}, the industrial application requires methods capable of localizing small or even minute defects on the surface of a 3D point cloud. To address this, we utilize the Fast Point Feature Histogram (FPFH), a manually defined point cloud feature descriptor, as the feature extractor for our feature-based method. By comparing the training and testing features obtained from the feature extractor, we can precisely localize defect areas. 

\subsubsection{Feature Extractor}
Inspired by BTF \citep{horwitz2022empirical}, our proposed method uses FPFH as the descriptor of local surface geometric features for industrial anomaly detection. FPFH is a manual point cloud feature descriptor \citep{FPFH}. Unlike neural network-extracted features, FPFH provides a robust representation of local geometric information with considerably lower computational cost \citep{Bergmann_Sattlegger_2023, wang2023multimodal, shapeguide}, which directly employs the relative coordinates and normal vector of each point.

To compute FPFH, we first calculate the Simplified Point Feature Histograms (SPFH), calculated according to Alg. \ref{Alg:1}. In Alg. \ref{Alg:1}, we use ``$\times$" to represent the cross-product. In the rest of the paper, ``$\times$" denotes the regular product unless otherwise specified. For a given point $a_i$ located on the point cloud and its nearby region $KNN(a_i)  \in \mathbb{R}^{1 \times m \times 3}$, built using k-nearest-neighbor (KNN) algorithm, we can calculate the tuple set for this point and its nearby region, $[\boldsymbol\alpha, \boldsymbol\gamma, \boldsymbol\theta]$, in Darboux frame ($\mu \nu w$) for every point pair $a_i$ and $a_j$ ($i \neq j$). The tuple set is then transferred as a histogram, i.e., $[\text{hist}(\boldsymbol\alpha), \text{hist}(\boldsymbol\gamma), \text{hist}(\boldsymbol\theta) ]$, which records the frequency of $[\boldsymbol\alpha, \boldsymbol\gamma, \boldsymbol\theta]$ in the $KNN(a_i)$. We denote SPFH of this point as $\phi_{SPFH}(a_i)$

\begin{figure}[!ht]
\centering\includegraphics[width=1.0\linewidth]{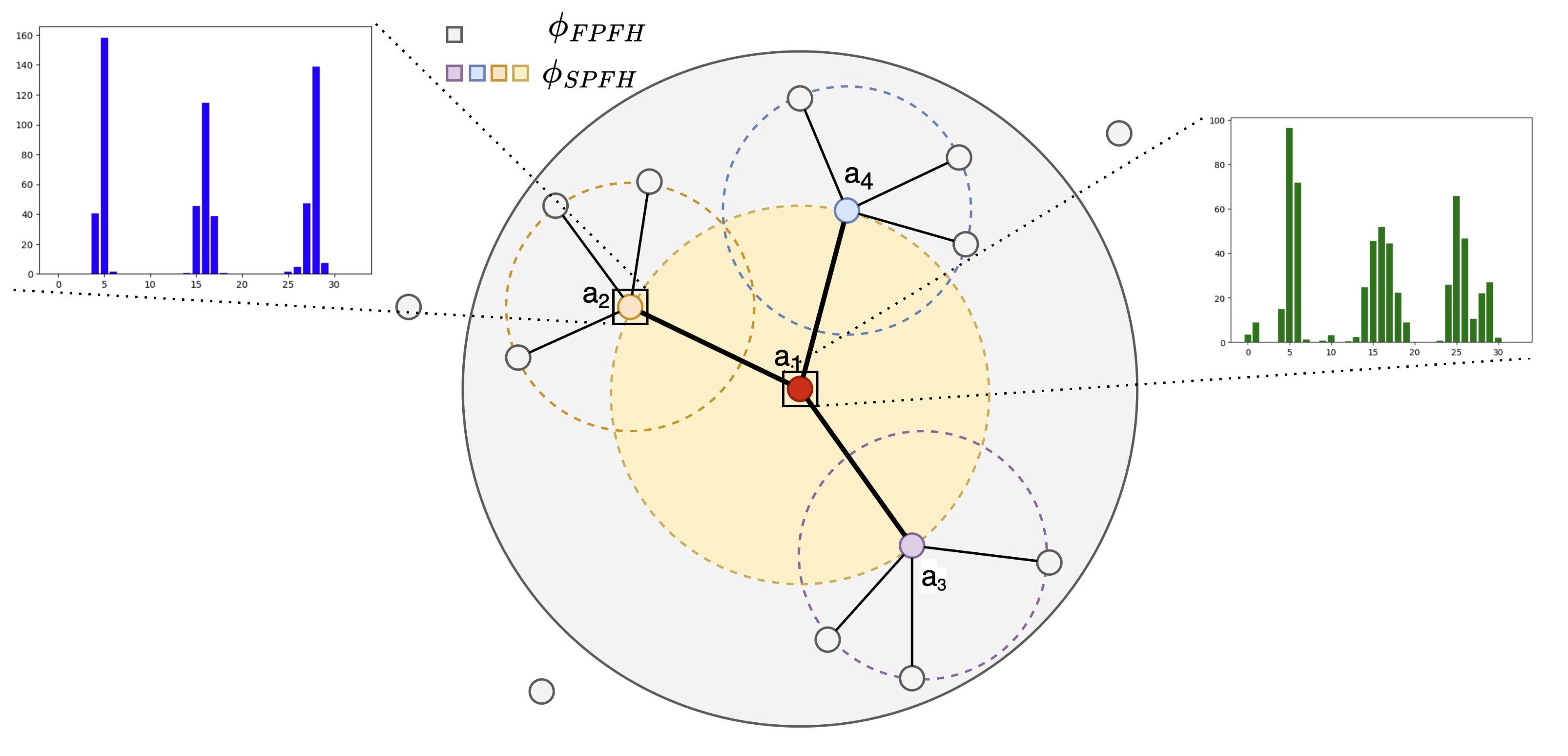}
\caption{The visual demonstration of SPFH and FPFH: each dashed circle represents the receptive field of SPFH of the center point, i.e., $a_{1}, a_2, a_3, a_4$, and the solid circle shows the receptive field of FPFH of $a_{1}$.}
\label{fig:SPFH}
\end{figure}

\begin{algorithm}[!ht]
\caption{Simplified Point Feature Histogram (SPFH)}\label{Alg:1}
\begin{algorithmic}
    \State{\textbf{Initiate}: KNN graph with $m$ points each $KNN(a_{i})$.  Denote $a_{i}$ the center point and $n_{i}$ is the normal of the center point.}
    \For{j = 1,2,...,m}
    
    \State{$\mu = n_{i}$, $\nu = \mu \times \frac{a_{j} - a_{i}}{||a_{j} - a_{i}||_{2}}$, $\omega = \mu \times \nu$}
    \State{$\alpha_{ij} = \nu \cdot n_{j}$, $\gamma_{ij} = \mu \cdot \frac{a_{j} - a_{i}}{||a_{j} - a_{i}||_{2}}$, $\theta_{ij} = \text{arctan}(\omega \cdot n_{j},\mu \cdot n_{j})$}
    \EndFor
    \State{$\boldsymbol\alpha_{i} = [\alpha_{i1},...,\alpha_{im}]$, $\boldsymbol\gamma_{i} = [\gamma_{i1},...,\gamma_{im}]$, $\boldsymbol\theta_{i} = [\theta_{i1},...,\theta_{im}]$}
    \State{
    $\phi_{SPFH}(a_{i})= [\text{hist}(\boldsymbol\alpha_{i}), \text{hist}(\boldsymbol\gamma_{i}), \text{hist}(\boldsymbol\theta_{i}) ]$
    } 
\end{algorithmic}
\end{algorithm}

Once the SPFH of all points on the 3D point cloud is computed, we can further compute the FPFH of the point cloud sample using Eq. (\ref{Eq:FPFH}), summarized in Alg. \ref{Alg:2}. Unlike SPFH, which is based on first-order nearest neighbor points, FPFH has a larger receptive field since it can aggregate the geometric information over both one-hop and two-hop neighbor points of the given point $a_{i}$. In addition, since FPFH adopts weights, $w = \frac{1}{||a_i - a_j||_2}$, it can be more accurate to depict the local geometric information than SPFH, defined as follows: 

\begin{equation}\label{Eq:FPFH}
    \phi_{FPFH}(a_i) = \phi_{SPFH}(a_i) + \frac{1}{m}\sum_{j=1}^{m} \frac{1}{||a_{i} - a_{j}||_{2}}\cdot \phi_{SPFH}(a_j).
\end{equation}


\begin{algorithm}[!ht]
\caption{Fast Point Feature Histogram (FPFH)}\label{Alg:2}
\begin{algorithmic}
    \State{\textbf{Input}: 3D Point Cloud Sample $P \in \mathbb{R}^{N \times 3}$.}
    \State{\textbf{Initiate}: $N$ KNN graphs with $m$ points each, $KNN(a) \in \mathbb{R}^{N \times m \times 3}$, denote $a$ as an arbitrary center point.}
    \For{i = 1,2,...,N}
   
    \State{
        $\phi_{SPFH}(a_{i})= [\text{hist}(\boldsymbol\alpha_{i}), \text{hist}(\boldsymbol\gamma_{i}), \text{hist}(\boldsymbol\theta_{i})]$
    }
    
    \EndFor  
    \For{i = 1,2,...,N}
    \For{j = 1,2,...,m}
    \State{
    $\phi_{FPFH}(a_i) = \phi_{SPFH}(a_i) + \frac{1}{m}\sum_{j=1}^{m} \frac{1}{||a_i-a_j||_{2}}\cdot \phi_{SPFH}(a_j)$
    }
    \EndFor
    \EndFor
\end{algorithmic}
\end{algorithm}

\subsubsection{PatchCore Anomaly Detection}
During training, the 3D point cloud $P \in \mathbb{R}^{N \times 3}$, containing $N$ points, is passed through the feature extractor and outputs FPFH features $\phi_{FPFH}(a_i)$ for each point. These features are then stored in the memory bank, i.e., $\mathcal{X}_M$. For convenience, we use $\phi$ to represent $\phi_{FPFH}$ for the remainder of the document. However, with a large number of training samples, the size of $\mathcal{X}_{M}$ becomes massive because the number of training samples determines the size of $\mathcal{X}_{M}$, leading to substantial memory usage and time-consuming inference. 
To mitigate this issue, we employ \textit{PatchCore} to reduce the inference time and achieve acceptable results simultaneously \citep{patchcore}. \textit{PatchCore} is a sub-sampling algorithm that constructs a Core-set $\mathcal{M}_c$. This algorithm adopts a minimax strategy, using a limited number of features to cover the entire feature space as much as possible \citep{sener2018active}. Hence, the Core-set $\mathcal{M}_c \in \mathcal{X}_M$ comprises the most representative features instead of saving all features. The number of representative features in the $\mathcal{M}_c$ is controlled by the pre-determined size $l$. Once the number of features in the $\mathcal{M}_c$ reaches the given size $l$, the algorithm stops sub-sampling. The algorithm is summarized in Alg. \ref{Alg:3}.

\begin{algorithm}[!ht]
\caption{FPFH Core-set Sampling}\label{Alg:3}
\begin{algorithmic}
    \State{\textbf{Input}: Memory bank of FPFH features $\mathcal{X}_M$, Core-set size $l$}
    \State{\textbf{Initiate:} $\mathcal{M}_c \xleftarrow{} \{ \}$, Randomly draw $ \phi(a_i) \in \mathcal{X}_M \xrightarrow{} \mathcal{M}_c$ }
    \While {$ \text{card}(\mathcal{M}_c) < l$}
    \State{$\phi(a_j) \leftarrow \underset{\phi(a_j) \in \mathcal{X}_M\setminus\mathcal{M}_C}{\arg \max } \underset{\phi(a_i) \in \mathcal{M}_C}{\min} \|\phi(a_i) -\phi(a_j)\|_2$}
    \State{$\mathcal{M}_c \leftarrow \mathcal{M}_c \cup \phi(a_j)$}
    \State{$\phi(a_i) = \phi(a_j)$}
    \EndWhile

    \State{\textbf{Return}: Core-set $\mathcal{M}_c$}
\end{algorithmic}
\end{algorithm}

\begin{figure}[!ht]
\centering\includegraphics[width=0.8\linewidth]{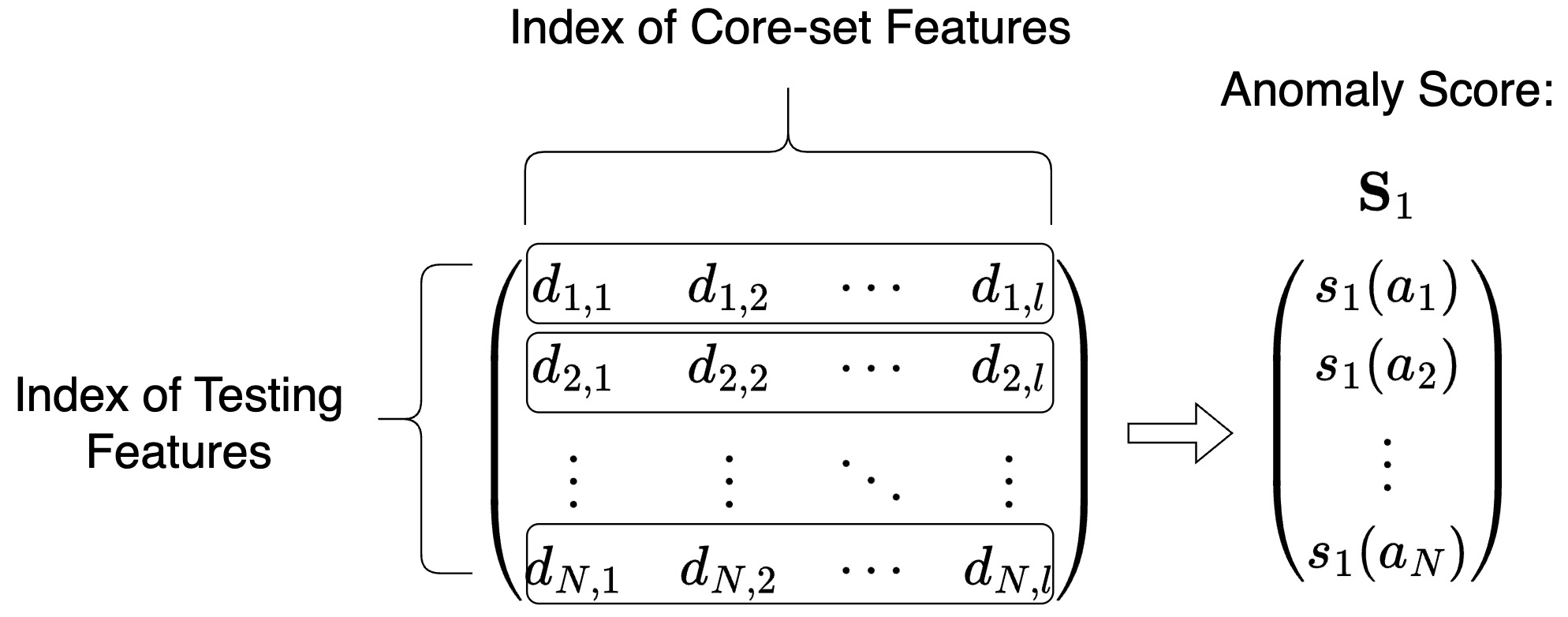}
\caption{Scheme of calculating the anomaly score $\textbf{S}_{1}$.}
\label{fig:s1}
\end{figure}

In the context of anomaly detection, the proposed module compares the features of the testing sample ${\phi}_{test}$ with the features in the $\mathcal{M}_c$ to yield a distance matrix, $\bold{D} \in \mathbb{R}^{N \times l}$. The row element for $\bold{D}$, denoted as $i$, indicates the distance of the $i_{th}$ testing features to all training features in $\mathcal{M}_c$.
To calculate the anomaly score for each sample, we extract the minimum value of each row, such that $\bold{S}_{1}: \mathbb{R}^{N \times l} \rightarrow \mathbb{R}^{N}$, where the suffix 1 indicates the first branch. The minimum value of each row represents the smallest distance of each testing feature to all training features in the Core-set. If the testing feature is the normal feature, this minimum value will be relatively small; conversely, a large value indicates a deviation from the normal features, suggesting defect points. 
The detailed algorithm and its implementation are depicted in Eq. (\ref{Eq:ano}) and Fig. \ref{fig:s1}.

\begin{equation}\label{Eq:ano}
\begin{split}
    d_{i,j} &= |\phi(a_{i}) - \phi(a_{j})|^{2}, i \in {1,...,N},\phi(a_{j}) \in \mathcal{M}_c \\
    s_{1}(a_i) &= \min_{j} (d_{i,j}), j = 1,..., l \\
    \bold{S}_1 &= [s_{1}(a_{1}),...,s_{1}(a_N)]
\end{split}
\end{equation}
where $\bold{S}_1$ is the anomaly score array for each sample.

\subsection{Reconstruction-based Anomaly Detection}\label{Sec:recon}
\begin{figure}[!ht]
\centering\includegraphics[width=1.0\linewidth]{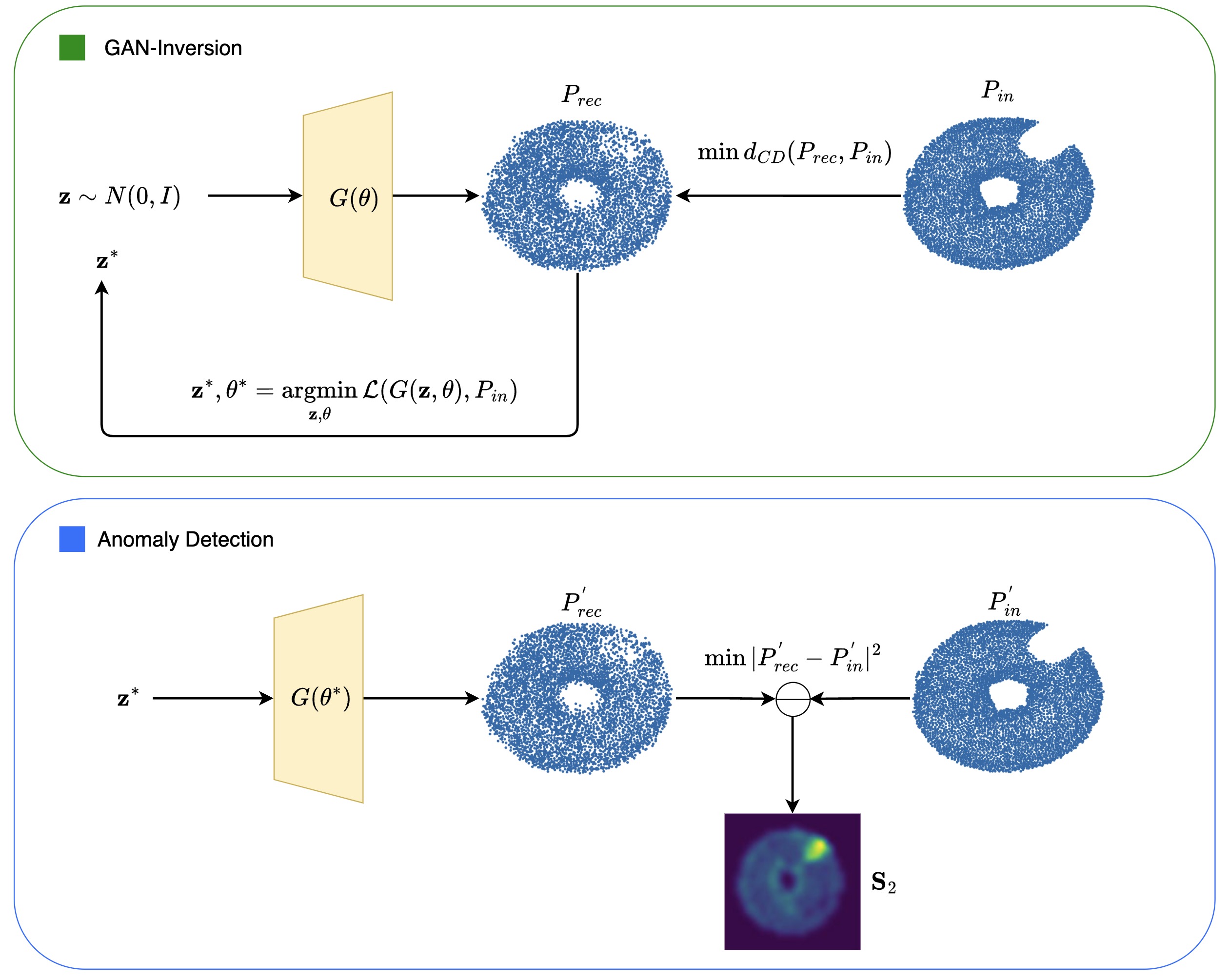}
\caption{The architecture of GAN-Inversion anomaly detection. There are two parts in our proposed module: (i) GAN-Inversion aims to find the latent code $\mathbf{z}^{*}$, and the network parameters $\theta^{*}$ can best reconstruct the testing samples; (ii) Anomaly Detection module involves an analytical comparison between the testing incomplete samples and the reconstructed samples, where the reconstructed samples are generated by the latent code $\textbf{z}^{*}$ and $\theta^{*}$ obtained GAN-Inversion module.}
\label{fig:gan}
\end{figure}

The feature-based detection module, introduced in Sec. \ref{sec:feature}, is effective at identifying local surface defects but fails to detect defects on the overall shape (i.e., missing regions), limiting the versatility of 3D anomaly detection. To overcome this limitation, we introduce a reconstruction-based method for identifying missing regions in the 3D point clouds by restoring the incomplete shapes. The design details are shown in Fig. \ref{fig:gan}. To the best of our knowledge, our proposed method is the first to adopt \textbf{GAN-Inversion} to a 3D anomaly detection task. In this section, we first introduce the mechanism of GAN and GAN-Inversion, followed by a discussion on how we apply GAN-Inversion to 3D anomaly detection. 

\subsubsection{Generative Adversarial Network (GAN)}
To identify the missing regions, we require a model that can represent anomaly-free samples based on GAN. 
GAN is a generative model \citep{gangoodfellow}, which is comprised of two foundational modules: a generator $G(\cdot;\theta)$ and a discriminator $D(\cdot;\theta')$, parameterized by $\theta$ and $\theta'$, respectively. We will refer to these functions as $G$ and $D$, respectively, when there is no ambiguity.
The generator aims to fool the discriminator by generating artificial samples, while the discriminator simultaneously attempts to distinguish the generated samples from real ones. When training is finished, the generator can be viewed as the data distribution of normal samples. Hence, the generated samples can be compared with testing samples to localize the missing regions. 
The training loss function of the GAN is denoted in Eq. (\ref{eq:ganloss}),

\begin{equation}
    \min _G \max _D V(D,G) = \mathbb{E}_{P\sim \mathcal{P}_{real}}[\log (D(P))]+\mathbb{E}_{\mathbf{z\sim \mathcal{N}(0,I)}} [\log (1-D(G(\mathbf{z})))],
\label{eq:ganloss}
\end{equation}
where $\mathcal{P}_{real}$ is the distribution of normal 3D point clouds.
The training stops when the loss in Eq. (\ref{eq:ganloss}) converges. After that, the model can draw a random $\mathbf{z} \in \mathcal{N}(0,I)$ and feed into the model using the following  Eq. (\ref{generator}), 

\begin{equation}
    P_{gen} = G(\mathbf{z};\theta).
\label{generator}
\end{equation}

This paper uses SP-GAN as a case study for the 3D point cloud generation \citep{li2021spgan}. Instead of turning a latent code $z$ into the 3D point cloud, SP-GAN adopts \textit{global prior} and \textit{local prior} to generate the 3D point clouds, where \textit{global prior}, provides general spatial guidance for the generation and \textit{local prior}, provides the surface details. Specifically, \textit{global prior} is the skeleton of the target point cloud, which fixes the number of points. In contrast, \textit{local prior} modifies the points on $S$ to minimize Eq. (\ref{eq:ganloss}) by the training of generator $G$ and discriminator $D$.

\subsubsection{GAN-Inversion}
In our experiment, GAN is trained on normal samples. Therefore, after training the GAN, it can be assumed that the model has learned the manifold of the normal products. However, for model-free products, variations in shape may occur even within the same type of product (i.e., no two apples are identical). Thus, a key question arises: How can we ensure the generated samples are always most similar to the input samples? 
Inspired by \textit{Shape-Inversion} \citep{shapeinversion}, we introduce GAN-inversion to ensure the generated sample closely resembles the input. The primary objective of this module is to always reconstruct the most similar normal sample to the input 3D point cloud, which is denoted as $P_{in} \in \mathbb{R}^{m \times 3}$, from a specific latent code $\textbf{z}^{*}$. This reconstruction is achieved by exploiting and finetuning the pre-trained GAN, as described in the following equation:

\begin{equation}
    \mathbf{z}^*=\underset{\mathbf{z} \in \mathbb{R}^d}{\arg \min } \mathcal{L}\left(G(\mathbf{z} ; \theta), P_{in}\right).
\label{ganinversion}
\end{equation}
This process is commonly referred to as Optimization-based GAN Inversion \citep{ganinversion}. Our model, however, extends beyond merely identifying the optimal latent code $\textbf{z}^{*}$ for reconstructing $P_{in}$. We simultaneously engage in the optimization of the GAN parameters $\theta$. This dual optimization process involves both the latent code and the GAN parameters and is essential for achieving a higher fidelity in the reconstructed 3D point cloud $P_{rec}$. This advanced method is shown in the Eq. (\ref{eq:ourinversion}):
\begin{equation}
    \theta^*, \mathbf{z}^*=\underset{\mathbf{z}, \theta}{\arg \min } \mathcal{L}\left(G(\mathbf{z} ; \theta), P_{i n}\right), \quad P_{rec}=G\left(\mathbf{z}^* ; \theta^{*}\right).
    \label{eq:ourinversion}
\end{equation}

In our study, the total loss function consists of Chamfer Distance and Feature Distance, shown in Eq. (\ref{eq:overall}):

\begin{equation}
    \mathcal{L} = \mathcal{L}_{CD} + \mathcal{L}_{FD}.
    \label{eq:overall}
\end{equation}

The detailed expression of $\mathcal{L}_{CD}$ is shown in Eq. (\ref{eq:CD}). The Chamfer distance mainly depicts the difference in the coordinates between two 3D point clouds. In Eq. (\ref{eq:CD}), $P_1$, $P_2$ are 3D point clouds, and $x$, $y$ are some points within the 3D point clouds. Chamfer distance is a metric to measure the similarity between 3D point clouds, which has two advantages: (i) Chamfer distance is a 2-norm distance, which has a low computational cost, and (ii) Chamfer distance is robust to outliers since it averages all the points. The Chamfer distance is defined as:

\begin{equation}
    \mathcal{L}_{C D}\left(P_1, P_2\right)=\frac{1}{\left|P_1\right|} \sum_{x \in P_1} \min _{y \in P_2}\|x-y\|_2+\frac{1}{\left|P_2\right|} \sum_{y \in P_2} \min _{x \in P_1}\|y-x\|_2.
    \label{eq:CD}
\end{equation}

Given that $\mathcal{L}_{CD}$ primarily addresses the positional attribute of the 3D point clouds, the Feature Distance, $\mathcal{L}_{FD}$, shown in Eq. (\ref{eq:feature dis}), is essential for maintaining the overall structural information:

\begin{equation}
    \mathcal{L}_{FD} = | D(P_{rec}) - D(P_{in}) |^{2},
    \label{eq:feature dis}
\end{equation}
where $D$ is the pre-trained discriminator

The inversion process starts from a latent code $\bold{z} \in \mathcal{N}(0,I)$. Then the model optimizes $\mathbf{z}$ and $\theta$ by minimizing the loss function in Eq. (\ref{eq:ourinversion}). When the inversion stage is completed, we assume that $P_{rec} = G(\mathbf{z}^{*};\theta^{*})$ is the closest non-defective sample of the input testing object.

\subsubsection{Incomplete Shape Detection}
Since we consider the missing region detection to be mainly determined by the shape of the normal product, the detailed geometric information on the local surface can be neglected when using reconstruction-based anomaly detection. Hence, we project reconstructed sample $P_{rec}$ and input testing sample $P_{in}$ on the 2D plane without $z$ coordinates, denoted as $P_{rec}^{'}$ and $P_{in}^{'}$. Using these two samples, these points in $P^{'}_{rec}$ while not in $P^{'}_{in}$ are likely from the missing regions, denoted as $\{p | p \in P^{'}_{rec}, p \notin P^{'}_{in} \}$, where $p$ is an arbitrary point. Using this definition, the anomaly score that indicates whether each point is from the missing region is defined in Eq. (\ref{eq:incompletion}):

\begin{equation}
    s_{2}(a_i) =  \min _{p \in P_{in}^{'}} \left|a_i - p\right|^{2}, a_i \in P_{rec}^{'}.
    \label{eq:incompletion}
\end{equation}

In this equation, the anomaly score is set as the distance from the point in the reconstructed sample to its closest point in the testing sample. In this way, the module assigns a small anomaly score for the points in the reconstructed sample with the same or nearby points on the testing sample. If there are missing regions in the testing sample, a high anomaly score will be assigned to points in the corresponding regions of the reconstructed sample (always be the closest normal sample to the testing one) as there is no nearby or same point in the testing sample. Once the computation is completed, we perform a density clustering algorithm to reduce the noise to get a robust result \citep{dbscan}. Since the anomaly regions are sparsely distributed, we dilate the defect regions for better results.

\subsection{Fusion Module}\label{sec:OCSVM}
So far, we have successfully adopted feature-based and reconstruction-based models for detecting defects on the local surface and incomplete shapes, respectively. However, in industrial settings, the conditions of the testing samples are often unknown. In other words, we do not have prior information about whether there are defects and what kinds of defects exist in the testing samples. To address this, we need to fuse the detection results from feature-based and reconstruction-based modules to produce an integrated anomaly score, which can uniformly identify all types of defects without any prior information.

\subsubsection{Anomaly Score Fusion}\label{sec:3.4.1}
In our model, each branch independently generates an anomaly score, which exhibits a unique scale. Hence, we propose OCSVM to fuse anomaly scores from different branches into an integrated anomaly score. OCSVM is an unsupervised anomaly detection method \citep{ocsvm}. Unlike traditional supervised support vector machines, OCSVM is only trained on single-class samples (normal samples in our case), making it well-suited for our unsupervised 3D anomaly detection task. The function of the decision boundary is defined in Eq. (\ref{Eq:decisionfunc}):

\begin{equation}\label{Eq:decisionfunc}
    f(\cdot) = \mathrm{sign}((w \cdot \Phi(\cdot) - \rho),
\end{equation}
where $w$ and $\rho$ are the optimized parameters and $\Phi$ is kernel function. 

\begin{figure}[!ht]
\centering\includegraphics[width=0.8\linewidth]{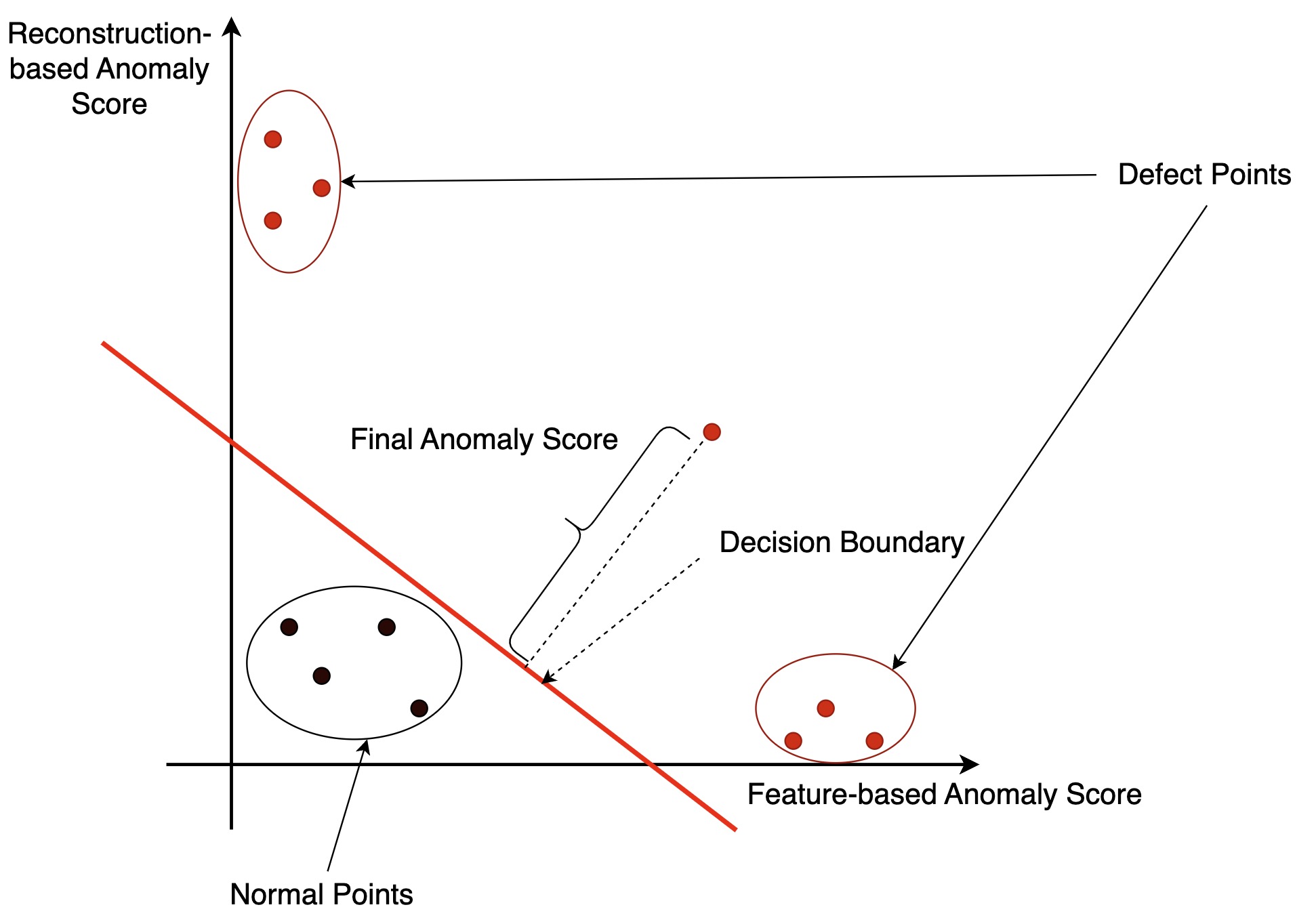}
\caption{Scheme of OCSVM, which computes the final anomaly score. Each point in this graph (red and black) represents the anomaly score pair.}
\label{fig:ocsvm}
\end{figure}

Given a 2D feature space, we define the horizontal axis as the anomaly score from the feature-based method and the vertical axis as the reconstruction-based anomaly score. Each point on the 2D space represents the joint anomaly score of a specific point on the point cloud. During training, we select and pass several samples from the training dataset through each branch to obtain two sets of anomaly scores, $\bold{S}_1$ and $\bold{S}_2$, for each point of the sample. We denote the anomaly score pairs as $(s_1(a_i), s_2(a_i))$. We then train our OCSVM and determine the decision boundary using these anomaly score pairs. Fig. \ref{fig:ocsvm} visually presents the mechanism of calculating the final anomaly score. 

To effectively train the decision boundary, it is necessary to adjust the anomaly scores from both branches to a comparable range. This involves finding the optimal coefficient $k$ between the two branches, such that 
$(s_1(a_i), s_2(a_i)) \mapsto (ks_1(a_i), s_2(a_i))$. In this paper, we adopt the ratio of the specific quantile of the anomaly scores from both branches as the coefficient $k$. Throughout the grid search, we obtain the optimal coefficient, $k^{*}$, such that $k^{*}$ allows the model to output the highest detection accuracy on the validation dataset (for details, see the ablation study in Sec. \ref{sec:4.3.2}) on the effect of different values of $k$). Once $k^{*}$ is determined, we follow the same training procedure during testing: passing the validation samples through each branch to acquire the joint anomaly scores in the 2D space, i.e., $ (k^{*} \times s_1(a_i), s_2(a_i))$. In this scenario, we define the final anomaly score as the distance of the joint anomaly score point to the decision boundary in the 2D feature space.

\section{Experiments}\label{sec:4}
\subsection{Experiments}
\subsubsection{dataset}
We evaluate our proposed method on the original MVTec 3D-AD dataset \citep{Bergmann_Jin_Sattlegger_Steger_2022b} and an augmented version of MVTec 3D-AD with additional samples that contain missing regions. A summary of the original MVTec 3D-AD dataset is shown in Tab. \ref{tab:orisum}. The dataset has ten categories, with only normal samples used for training. It is worth noticing that the dataset contains multiple categories of model-free products, such as ``Bagel", ``Cookie", ``Peach", etc. The testing dataset includes both anomaly-free and anomalous samples, such as scratches, dents, holes, contamination, or combinations of multiple defects. In Tab. \ref{tab:orisum}, we use ``Test (good)" and ``Test" to represent anomaly-free and anomalous samples in testing, respectively. As discussed in Sec. \ref{sec:3.4.1}, we conduct a grid search to find the optimal coefficient $k$ for each category for determining the decision boundary. To facilitate this, we create a validation dataset comprising 5 samples of each defect type for every product category, along with 5 anomaly-free samples to maintain a distribution consistent with the testing dataset. 

The augmented version of the MVTec 3D-AD dataset is created by adding abnormal samples with missing regions for the ``Bagel", ``Peach, and ``Potato" categories while keeping the remaining categories unchanged, as summarized in Tab. \ref{Tab:augsum}. It is worth noticing that the ``Cookie" category in the original dataset already contains several incomplete shape samples in the original test set. 

\begin{table}[!ht]
\caption{Summary of Custom Dataset \citep{Bergmann_Jin_Sattlegger_Steger_2022b}.}\label{tab:orisum}
\centering
{\fontsize{10}{14}\selectfont
\begin{tabular}{lccccc}
\toprule
Category & \# Train & \# Test (good) & \# Test & \# Validation & \# Defect types \\ 
\midrule
bagel & 244 & 17 & 68 & 25 & 4 \\
cable gland & 223 & 16 & 67 & 25 & 4 \\
carrot & 286 & 22 & 107 & 30 & 5  \\
cookie & 210 & 23 & 83 & 25 & 4 \\
dowel & 288 & 21 & 84 & 25 & 4 \\
foam & 236 & 15 & 60 & 25& 4 \\
peach & 361 & 21 & 86 & 25 & 4 \\
potato & 300 & 17 & 72 & 25& 4 \\
rope & 298 & 27 & 54 & 20 & 3 \\
tire & 210 & 20 & 67 & 25 & 4 \\
\midrule
total & 2656 & 199 & 748 &  250 & 40 \\
\bottomrule
\end{tabular}
}
\end{table}

\begin{table}[!ht]
\caption{Summary of Augmented Dataset.}\label{Tab:augsum}
\centering
{\fontsize{10}{14}\selectfont
\begin{tabular}{lcccccc}
\toprule
Category & \# Training & \# Testing (total) & \# Testing (incomplete) & \# Validation \\ 
& & & \\
\midrule
bagel & 244 & 102 & 17 & 30 \\
peach & 361 & 144 & 37 & 30\\
potato & 300 & 112 & 28 & 35 \\
\bottomrule
\end{tabular}
}
\end{table}

\subsubsection{Data Preprocessing}
For data preprocessing, we follow the methods described in BTF \citep{horwitz2022empirical}. Initially, we use the RANSAC algorithm to estimate the background plane and remove the background in the training and testing dataset, thereby enhancing the computation speed. Subsequently, we downsample the positional tensors of the entire dataset to a resolution of $224 \times 224$ to further conserve computation. 
We implement our experiments on NVIDIA A5000 GPUs using PyTorch \citep{NEURIPS2019_9015}.

\subsubsection{Parameter Settings}
In the feature-based module, we select 30 neighbor points to construct the $KNN$ graph. We then $l = 0.01$ to build the memory bank $\mathcal{X}_{m}$ and Core-set $\mathcal{M}_c$. 

We train our base GAN model across all categories using the latent code $\bold{z} \in \mathbb{R}^{128}$ and output the reconstructed point cloud $P_{rec} \in \mathbb{R}^{4096 \times 3}$, which has $4,096$ points. The training process was set to 300 epochs. In the GAN-Inversion component, we use a two-stage optimization strategy for faster and more accurate reconstruction of the incoming testing samples. The learning rate $\alpha$ for the latent code $\bold{z}$ and $\theta$ is set to $(\alpha_{\bold{z}}, \alpha_{\theta}) = (2 \times 10^{-5},1 \times 10^{-4})$ for the first stage and $(\alpha_{\bold{z}}, \alpha_{\theta}) = (1 \times 10^{-5},1 \times 10^{-5})$ for the second stage. Each stage consists of 40 iterations.

\subsubsection{Evaluation Metric}\label{sec:4.1.4}
Our study focuses on pixel-level anomaly detection tasks. Hence, we follow the Area Under Per-Region Overlap (AU-PRO) metric introduced in \citep{Bergmann_Jin_Sattlegger_Steger_2022b}. 

\begin{equation}
    \mathrm{AU} \text{-}\mathrm{PRO} =\frac{1}{K} \sum_{k=1}^K \frac{\left|Q \cap C_k\right|}{\left|C_k\right|}
\end{equation}

\begin{figure}[!ht]
\centering\includegraphics[width=0.6\linewidth]{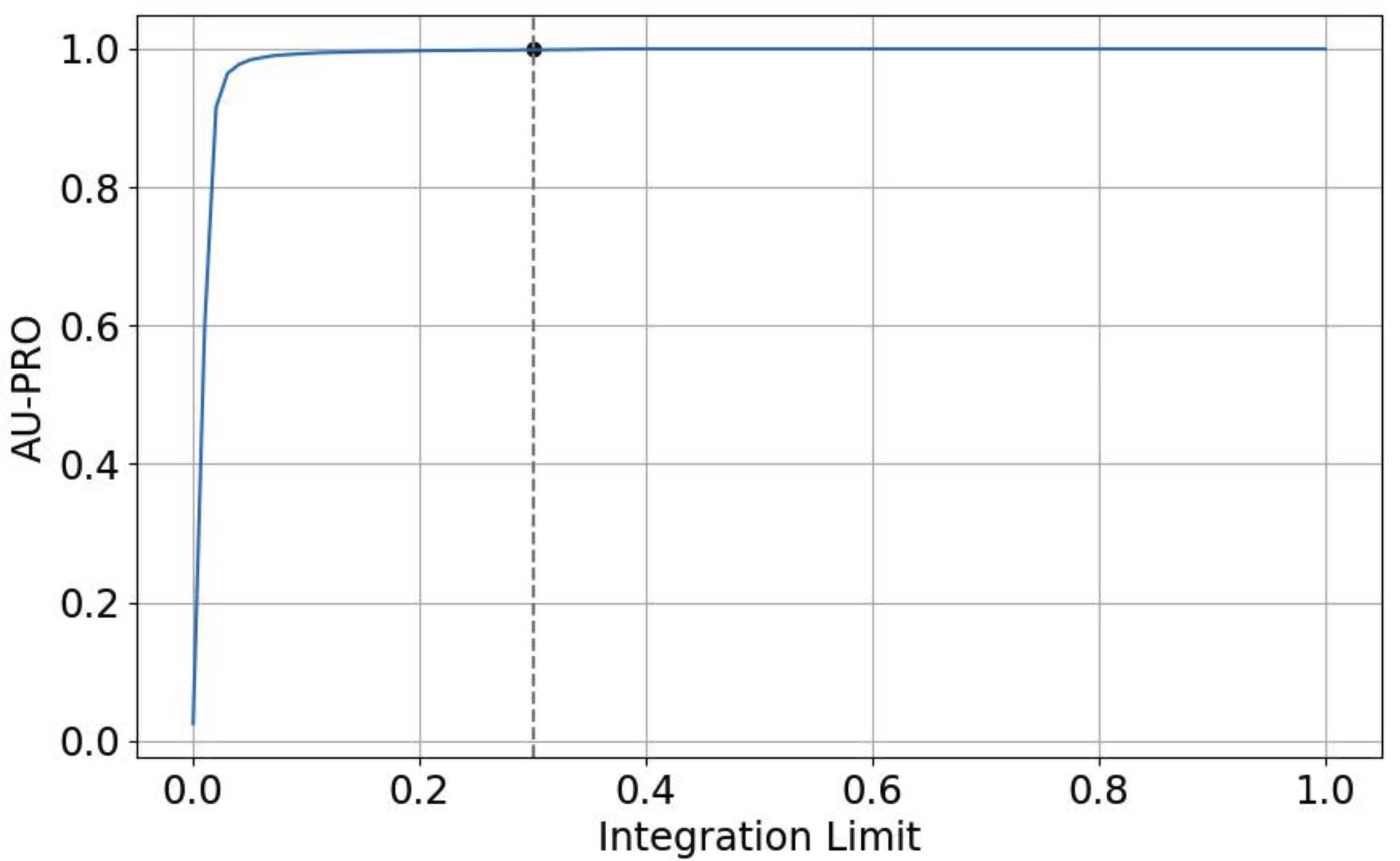}
\caption{The scheme of AU-PRO. The reported value is the area under the \textcolor{RoyalBlue}{blue} curve from 0.0 to 0.3.}
\label{fig:aupro}
\end{figure}

The AU-PRO metric ranges from 0 to 1, in which a higher score means better model performance. In this study, we set $0.3$ as the integration limit (false positive rate) to ensure a fair comparison with other benchmark methods.  As illustrated in Fig. \ref{fig:aupro}, the AU-PRO is directly proportional to the integration limit. The value of $0.3$ is considered an ideal value for industrial anomaly detection tasks, where the abnormal regions are typically tiny compared to normal regions \citep{Bergmann_Jin_Sattlegger_Steger_2022b}. A high AU-PRO at a given integration limit reflects the robust performance of the model.

\subsection{Results}\label{sec:4.2}
Tab. \ref{tab:aupro1} presents the performance of the proposed method, measured by AU-PRO, alongside that of current SOTA methods on the original and augmented datasets. We compare our method with BTF \citep{horwitz2022empirical}, M3DM \citep{wang2023multimodal}, and Shape-guided method \citep{shapeguide}. On the original dataset, our method achieves the best performance in anomaly detection across five out of the ten product categories and also the highest overall performance. It is worth noting that in the original dataset, ``Cookie" is the only category of the product containing the missing region. Our proposed method outperforms all benchmark methods by a large margin (improving over the second-best method by 2.6 in AU-PRO). To further demonstrate the superior performance of our proposed method in identifying the missing regions, we also evaluate it on the augmented dataset. Specifically, our proposed method largely outperforms the benchmarks in the categories of ``Bagel", ``Peach", and ``Potato", in which we manually added samples with missing regions. The results clearly demonstrate that the benchmark methods apparently failed in anomaly detection when a large number of samples have missing regions, while our proposed method still maintains outstanding performance and significantly outperforms the existing benchmarks.

Incomplete shape detection is critical in industrial anomaly detection, for which current methods fail to localize. With the reconstruction-based method and OCSVM, it is possible to detect missing regions on the model-free products, such as ``Bagel", ``Cookie", ``Peach", ``Potato", etc., for subsequent restoration.

\begin{table}[!ht]
\centering
\caption{The anomaly localization performance evaluated by AU-PRO on the original MVTec 3D-AD Dataset (Top) and the augmented dataset (Bottom). The best results are marked in bold.}
\label{tab:aupro1}
{\fontsize{7.5}{14}\selectfont  
\begin{tabularx}{\textwidth}{l*1|l*{10}{X}c}
\toprule
Methods & Bagel & Cable Gland & Carrot & Cookie & Dowel & Foam & Peach & Potato & Rope & Tire & Mean \\
\midrule
BTF (FPFH) & \textbf{97.3} & 87.1 & \textbf{98.2} & 90.7 & 88.8& 74.2 & 97.7 & 98.2 & \textbf{95.5} & 96.2 & 92.4 \\
M3DM (3D) & 94.4 & 81.6 & 97.5 & 88.4 & 87.7 & \textbf{77.6} & 95.6 & 97.5 & 94.7 & 93.8 & 90.9 \\
Shape-guided method (3D) & 97.2 & 85.2 & 98.1 & 92.1 & \textbf{89.5} & 76.2 & \textbf{97.8} & \textbf{98.3} & 95.4 & \textbf{97} & 92.7 \\
Ours & \textbf{97.3} & \textbf{87.4} & \textbf{98.2} & \textbf{94.7} & 88.8 & 76.6 & 97.7 & 98.2 & \textbf{95.5} & 96.3 & \textbf{93.0} \\
\bottomrule
\end{tabularx}
}
\\[12pt]

{\fontsize{7.5}{14}\selectfont  
\begin{tabularx}{\textwidth}{l*1|l*{10}{X}c}
\toprule
Methods & Bagel & Cable Gland & Carrot & Cookie & Dowel & Foam & Peach & Potato & Rope & Tire & Mean \\
\midrule
BTF (FPFH) & 83.3 & 87.1 & \textbf{98.2} & 90.7 & 88.8 & 74.2 & 80.1 & 87.4& \textbf{95.5} & 96.2  & 88.2 \\
M3DM (3D) & 81.9 & 81.6 & 97.5 & 88.4 & 87.7 & \textbf{77.6} &77.4 & 84.9  & 94.7 & 93.8 & 86.6 \\
Shape-guided method (3D) & 82.5 & 85.2 & 98.1 & 92.1 & \textbf{89.5} & 76.2  & 77.1& 80.8 & 95.4 & \textbf{96.9} & 87.4 \\
Ours & \textbf{97.3} & \textbf{87.4} & \textbf{98.2} & \textbf{94.7} & 88.8 & 76.6 &\textbf{97.5} & \textbf{95.6} & \textbf{95.5} & 96.3 &\textbf{92.8} \\
\bottomrule
\end{tabularx}
}
\end{table}

\subsection{Ablation Study}
We conduct the following three ablation experiments to better understand our proposed method: (i) examining the compatibility of different feature-based methods with the reconstruction-based method using OCSVM as the fusion module; (ii) discussing the impact of choosing different $k^{*}$ on the accuracy when training the decision boundary for OCSVM; (iii) comparing the detection results of different fusion methods compared with OCSVM.

\subsubsection{Compatibility of GAN-Inversion Module}
We evaluate the compatibility of our GAN-Inversion anomaly detection module by embedding it with the Shape-guided method and M3DM. 
Tab. \ref{tab:aupro2} shows the AU-PRO accuracy for both the original and the augmented datasets. The results indicate that the Shape-guided method + GAN maintains comparable accuracy on the original dataset and improves the accuracy of the augmented dataset. On the other hand, although M3DM + GAN effectively improves the detection accuracy on the augmented dataset, the overall accuracy drops 0.5 percentage points on the original dataset. 

In conclusion, the results of Tab. \ref{tab:aupro2} suggest that the reconstruction and fusion modules are versatile, plug-and-play modules that can be effectively integrated with current SOTA feature-based methods. Additionally, these plug-and-play modules improve upon the vanilla SOTA detection accuracy, particularly in categories with incomplete samples. Although the reconstruction-based and fusion modules generally improve the detection accuracy of incomplete shapes, using BTF proves to be the best fit for the feature-based module.

\begin{table}[!ht]
\centering
\caption{The performance of benchmarks with GAN-Inversion module (AU-PRO) on the original MVTec 3D-AD Dataset (Top) and the augmented dataset (Bottom). The best results are marked in bold.}
\label{tab:aupro2}
{\fontsize{7.5}{14}\selectfont  
\begin{tabularx}{\textwidth}{l*1|l*{10}{X}c}
\toprule
Methods & Bagel & Cable Gland & Carrot & Cookie & Dowel & Foam & Peach & Potato & Rope & Tire & Mean \\
\midrule
M3DM (3D) & 94.4 & 81.6 & 97.5 & 88.4 & 87.7 & \textbf{77.6} & 95.6 & 97.5 & 94.7 & 93.8 & 90.9 \\
M3DM + GAN & 95.3 & 79.1 & 97 & 92.7 & 84.5 & 74.4 & 96.3 & 97.6 & 93.9 & 90.7 & 90.2 \\
\midrule
Shape-guided method (3D) & \textbf{97.2} & \textbf{85.2} & \textbf{98.1} & 92.1 & \textbf{89.5} & 76.2 & \textbf{97.8} & \textbf{98.3} & \textbf{95.4} & \textbf{96.9} & 92.7 \\
Shape-guided method + GAN &  96.8 & 85.3 & 98 & \textbf{96.1} & 87.6 & 76.2 & 97.5 & \textbf{98.3} & 95.3 & 96.8 & \textbf{92.8} \\
\bottomrule
\end{tabularx}
}
\\[12pt]
{\fontsize{7.5}{14}\selectfont  
\begin{tabularx}{\textwidth}{l*1|l*{10}{X}c}
\toprule
Methods & Bagel & Cable Gland & Carrot & Cookie & Dowel & Foam & Peach & Potato & Rope & Tire & Mean \\
\midrule
M3DM (3D) & 81.9 & 81.6 & 97.5 & 88.4 & 87.7 & \textbf{77.6} &77.4 & 84.9  & 94.7 & 93.8 & 86.6 \\
M3DM + GAN & 94.1 & 79.1 & 97 & 92.7& 84.5& 74.4& 94.8 & 95.2 &93.9 &90.7& 89.6 \\
\midrule
Shape-guided method (3D) & 82.5 & \textbf{85.2} & \textbf{98.1} & 92.1 & \textbf{89.5} & 76.2  & 77.1& 80.8 & \textbf{95.4} & \textbf{96.9} & 87.4 \\
Shape-guided method + GAN & \textbf{97} & 85.3& 98& \textbf{96.1}& 87.6&76.2&\textbf{97.6} & \textbf{97.1} &95.3& 96.8& \textbf{92.7}\\
\bottomrule
\end{tabularx}
}
\end{table}

\subsubsection{Impact of Coefficient $k$}\label{sec:4.3.2}

In this subsection, we investigate the impacts of the coefficient $k$ on the accuracy of the validation dataset. Our proposed method requires an adjustable ratio $k$ to effectively scale anomaly scores from the two distinct branches of anomaly detection into a comparable range. This is essential since each branch generates anomaly scores with different scales. Ensuring that anomaly scores from both branches are balanced is crucial to avoid any single branch disproportionately influencing the final anomaly score, thereby ensuring effective detection of all defect types.

\begin{figure}[!ht]
\centering\includegraphics[width=0.7\linewidth]{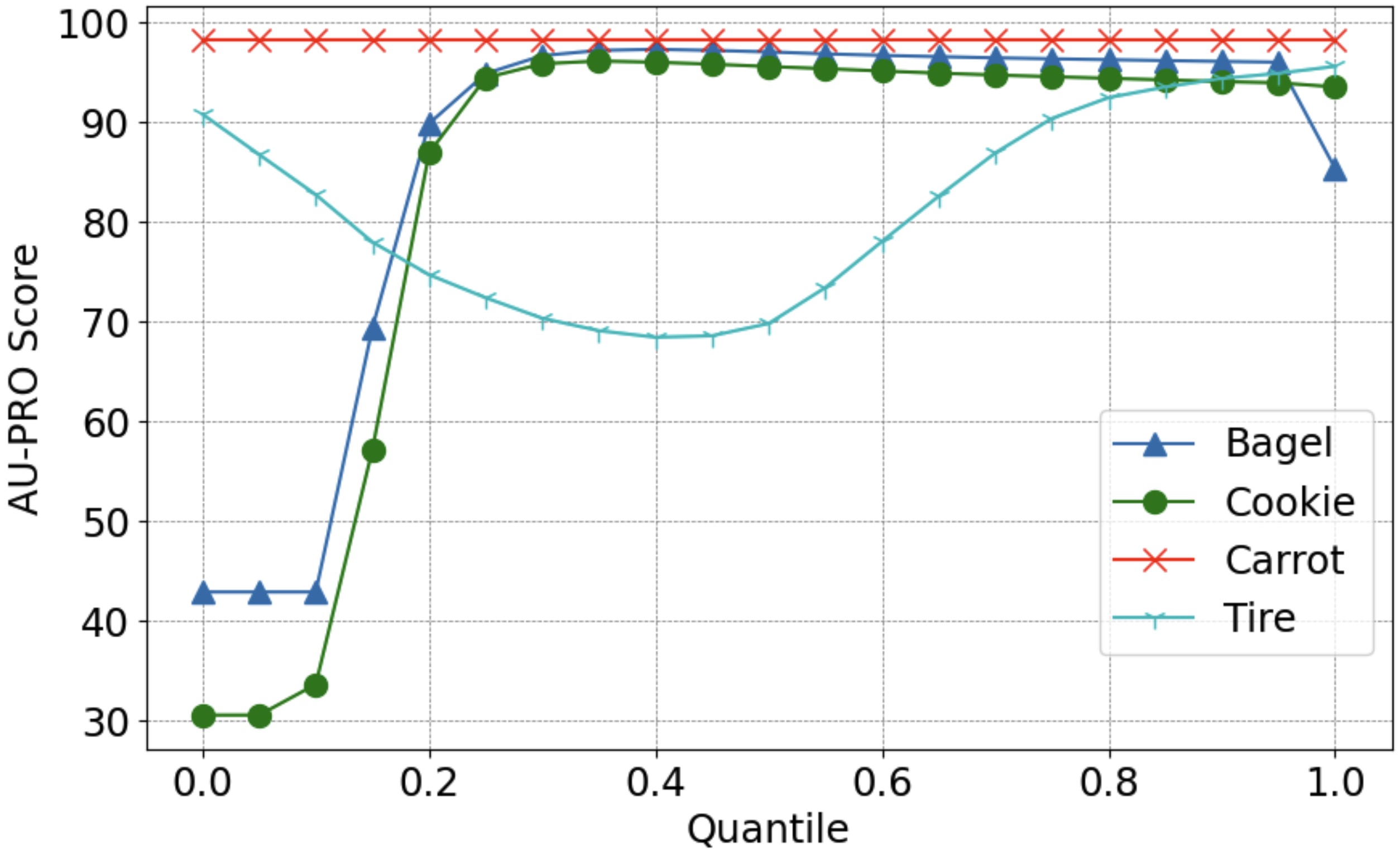}
        \caption{The evaluation metric of AU-PRO on the augmented dataset of different ratios. We report AU-PRO of 4 categories on the validation set: Bagel, Cookie, Carrot, and Tire.}
\label{fig:ablationfix}
\end{figure}

In Fig. \ref{fig:ablationfix}, we plot AU-PRO on the validation dataset versus the ratio that is adjusted by $k$ (with values from 0.0 to 1 quantile). To determine the optimal $k$, we perform a grid search across the given range of quantiles, where we choose 21 ratios and calculate the accuracy on the validation set for each setting. The model then selects the $k$ with the highest accuracy to carry forward into the subsequent inference stage. The results demonstrate that selecting an appropriate coefficient $k$ can largely improve the detection accuracy for each category. 

\subsubsection{Contribution of OCSVM}
In this subsection, we evaluate the effectiveness of the OCSVM module. We compare our proposed method, the benchmark (BTF), and the fusion method (using weight ($w$) and bias ($b$)) introduced in the Shape-guided method \citep{shapeguide}. In Shape-guided method, \cite{shapeguide} adopt  Eq. (\ref{eq:13}) to adjust two anomaly scores: 

\begin{equation}\label{eq:13}
\begin{split}
    &w = \frac{\sigma_1}{\sigma_2} \\
    &b = \mu_{1} - w \cdot \mu_{2} \\
    &\bold{S}_2^{'} = w \cdot \bold{S}_2+ b \\
    &\bold{S}_{final} = \bold{S}_1 + \bold{S}_2^{'}
\end{split}
\end{equation}
where $\sigma_i$ and $\mu_i$ are the standard deviation and mean of anomaly score $\bold{S}_i$ from the $i_\text{th}$ branch on the validation set. 
The idea of Eq. (\ref{eq:13}) is also to adjust the anomaly scores from one branch to be compatible with another using the validation set. The Shape-guided method views anomaly scores from each branch as a specific distribution. By aligning the means ($\mu$) of two anomaly scores using the bias ($b$) and scaling the range with the weight ($w$), the method computes the scaled anomaly score $\bold{S}_{2}^{'}$ for one branch. The final anomaly score is then calculated as the summation of two anomaly scores. In this design, the fusion module only accesses the validation set once to obtain $w$ and $b$, which avoids the grid search in our proposed method and reduces the inference time. However, Tab. \ref{tab:contribution} shows that  OCSVM outperforms the fusion method using mean and standard deviation on all categories for the original dataset, which demonstrates the effectiveness of the proposed fusion module.

\begin{table}[!ht]
\centering
\caption{Contribution of OCSVM on the original dataset.}
\label{tab:contribution}
{\fontsize{7.5}{14}\selectfont  
\begin{tabularx}{\textwidth}{l*1|l*{10}{X}c}
\toprule
Methods & Bagel & Cable Gland & Carrot & Cookie & Dowel & Foam & Peach & Potato & Rope & Tire & Mean \\
\midrule
BTF (FPFH) & \textbf{97.3} & 87.1 & \textbf{98.2} & 90.7 & \textbf{88.8}& 74.2 & \textbf{97.7} & \textbf{98.2} & \textbf{95.5} & 96.2 & 92.4 \\
BTF+GAN & 95 & 81.8 & 96.6 & 91.7 & 84.2 & 76.2 & 95.9 & 98.1 & 94.1 & 80.6 & 89.4\\
Ours$_{OCSVM}$ & \textbf{97.3} & \textbf{87.4} & \textbf{98.2} & \textbf{94.7} & \textbf{88.8} & \textbf{76.6} & \textbf{97.7} & \textbf{98.2} & \textbf{95.5} & \textbf{96.3} & \textbf{93.0} \\
\bottomrule
\end{tabularx}
}
\end{table}

\section{Conclusion}\label{sec:5}
This paper addresses the research gap concerning the lack of a unified 3D anomaly detection method that can handle all types of defects in model-free products with complex geometric features. Specifically, feature-based methods struggle to identify anomalies in incomplete shapes, while reconstruction-based approaches have not yet achieved the required resolution for the detection of tiny defects for industrial standards. No method has been proposed to unify both of these methods to overcome the weaknesses present in these individual methods.

To address these problems, we proposed a unified framework for unsupervised detection of universal industrial product defects. To our knowledge, this is the first method to adopt GAN-inversion for the 3D point clouds anomaly detection task. Our method has three critical modules: feature-based, reconstruction-based, and fusion. The feature-based module is specifically designed to identify geometric defects on product surfaces, while the reconstruction-based module evaluates the completeness of product shapes. In the last step, we employ OCSVM to integrate the results from the two modules and generate a final anomaly score. To evaluate our method, we create an augmented dataset with artificially generated incomplete samples, given the absence of such samples in the original dataset. Our experimental results in Tab. \ref{tab:aupro1} show that our method not only performs comparably with existing SOTA methods on the original dataset but also significantly outperforms them on the augmented dataset. This demonstrates the robustness and adaptability of our proposed method. Overall, the advantages of our proposed method can be summarized in two key aspects: (i) our proposed method is an unsupervised 3D anomaly detection for model-free products; (ii) it uses the feature-based method, reconstruction-based method and a fusion module, which improves the detection accuracy across defect types. 


\section*{Acknowledgements}
This research is partially supported by the Air Force Office of Scientific Research (AFOSR) under the Dynamic Data and Information Processing (DDIP) portfolio via Grant  23AFCOR003.   


\nocite{*} 

\bibliography{sample} 



\end{document}